\documentclass[letterpaper]{article} % DO NOT CHANGE THIS
\usepackage{aaai2026}  % DO NOT CHANGE THIS
\usepackage{times}  % DO NOT CHANGE THIS
\usepackage{helvet}  % DO NOT CHANGE THIS
\usepackage{courier}  % DO NOT CHANGE THIS
\usepackage[hyphens]{url}  % DO NOT CHANGE THIS
\usepackage{graphicx} % DO NOT CHANGE THIS
\usepackage{natbib}  % DO NOT CHANGE THIS AND DO NOT ADD ANY OPTIONS TO IT
\usepackage{caption} % DO NOT CHANGE THIS AND DO NOT ADD ANY OPTIONS TO IT
\usepackage[utf8]{inputenc} % allow utf-8 input
\usepackage[T1]{fontenc}    % use 8-bit T1 fonts

\usepackage{booktabs}       % professional-quality tables
\usepackage{amsfonts}       % blackboard math symbols
\usepackage{nicefrac}       % compact symbols for 1/2, etc.
\usepackage{microtype}      % microtypography
\usepackage{amsmath}

\usepackage{array}
\usepackage{pifont}
\usepackage{enumitem}

\usepackage{multirow}
\usepackage{colortbl}
\graphicspath{{Figure/}}
\newcommand{\cmark}{\ding{51}}
\newcommand{\xmark}{\ding{55}}

\nocopyright

\title{Customizing Video Portraits via Identity-Action Decoupling}
\author{
    Junxiong Lin\textsuperscript{\rm 1},
    Haoran Wang\textsuperscript{\rm 1},
    Xinji Mai\textsuperscript{\rm 1},
    Zeng Tao\textsuperscript{\rm 1},\\
    Xuan Tong\textsuperscript{\rm 1},
    Ivy Pan\textsuperscript{\rm 3},
    Wenqiang Zhang\textsuperscript{\rm 1,\rm 2}
}
\affiliations{
    \textsuperscript{\rm 1}College of Intelligent Robotics and Advanced Manufacturing, Fudan University\\
    \textsuperscript{\rm 2}College of Computer Science and Artificial Intelligence, Fudan University\\
    \textsuperscript{\rm 3}The University of Hong Kong
}

\begin{document}

\maketitle

% ---- figure ----  
\begin{figure*}[h]
  \centering
  % \fbox{\rule{0pt}{2in} \rule{0.97\linewidth}{0pt}}
    \includegraphics[width=1\linewidth]{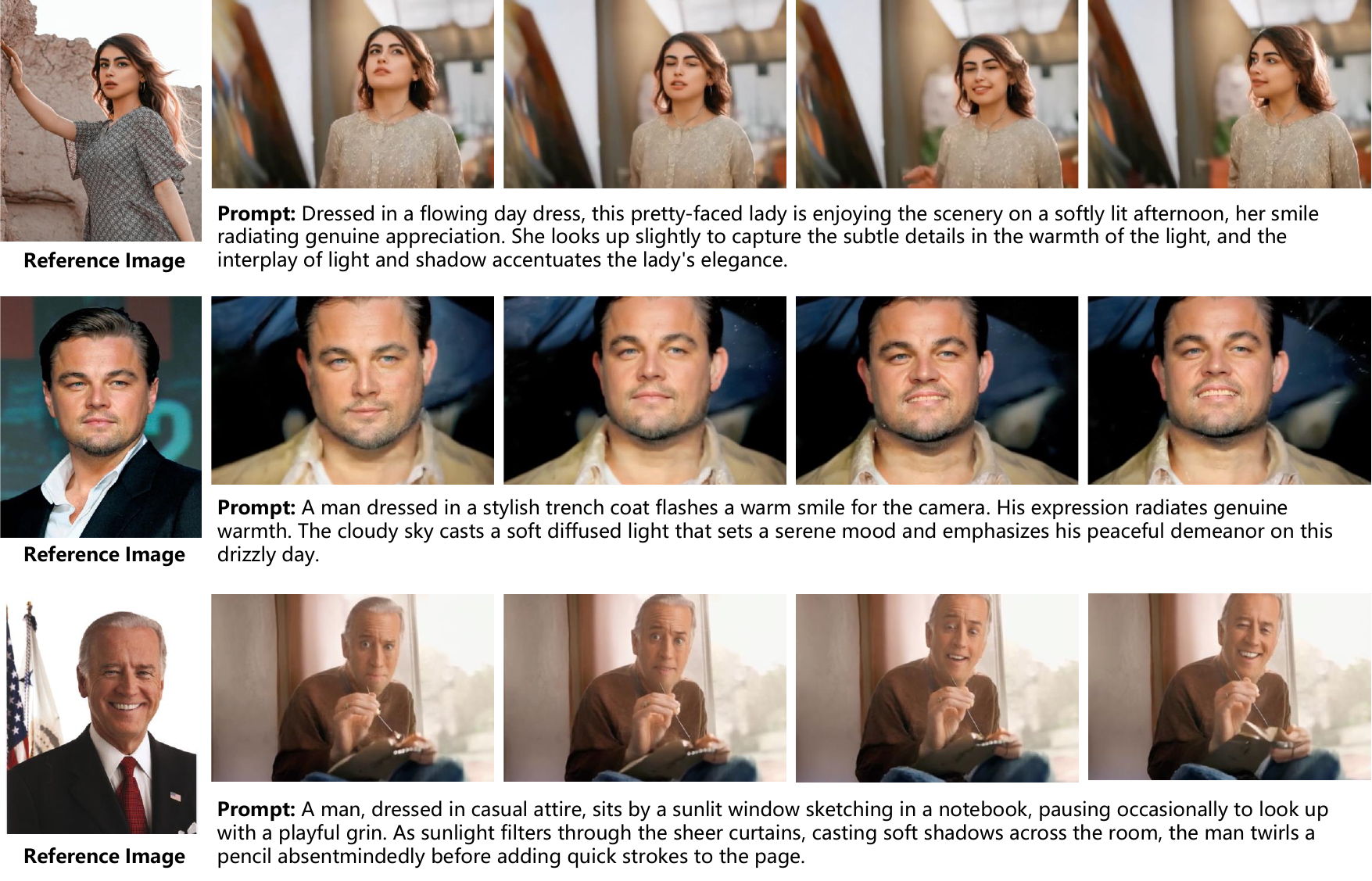}
   \caption{Examples of identity-preserving video generation (IPT2V) by our IaD. Given a reference image, IaD can generate customized videos based on the text prompt while preserving the identity of reference image.}
   \label{fig:top}
\end{figure*}
% ---- ---- ---- ---- ----

\begin{abstract}
Identity-Preserving Text-to-Video Generation (IPT2V) seeks to synthesize a temporally coherent video from a reference image and a textual description, while simultaneously preserving the subject’s identity and allowing fine-grained control over facial dynamics. Although recent methods such as ID-Animator and ConsisID inject identity features only at inference time, they ignored the ID-irrelevant information contained in Facial embedding, leading to monotonous or inaccurate facial movements that poorly follow the prompt. We introduce Identity–Action Decoupling (IaD) framework as well as two loss function Identity Decoupling Loss $L_{ID}$ and Text Alignment Loss $L_{TA}$ to solve this problem. Without any subject-specific fine-tuning, IaD yields videos that (1) maintain cross-temporal identity consistency and (2) exhibit rich, controllable expressions and scene variations that closely match the input text. 
\end{abstract}

\section{Introduction}
Identity-Preserving Text-to-Video Generation (IPT2V) seeks to synthesize video sequences conditioned on a single reference image and an accompanying textual description, while simultaneously preserving the subject’s identity across all frames and enabling fine-grained control over facial dynamics \cite{chefer2024still,yuan2024identity,kim2025subject,he2024id}.
By simultaneously ensuring cross-temporal identity consistency and enabling fine-grained facial control, IPT2V holds substantial potential for a wide range of applications, such as personalized digital humans and immersive virtual try-on systems \cite{song2024idprotector, li2024personalvideo,wang2024lia,tong2025component,wang2026hi,tao2024lcgen,qiu2024moviecharacter}.

Benefiting from the expressive power of large-scale, pre-trained video diffusion models and parameter-efficient adaptation techniques such as LoRA \cite{hu2022lora} and Textual Inversion \cite{gal2022image}, recent approaches such as Still-Moving \cite{chefer2024still} can already synthesize high-quality, personalized videos after fine-tuning on only a handful of user-provided reference images. However, these methods continue to depend on subject-specific training at test time, rendering the deployment pipeline both cumbersome and computationally expensive \cite{fei2025ingredients,deng2025cinema,mai2026agentic,mai2025cues}. As a result, research is shifting away from per-case fine-tuning toward unified, subject-agnostic adaptation strategies: for instance, ID-Animator \cite{he2024id} and ConsisID \cite{yuan2024identity} inject identity features solely during inference through lightweight modifications to the diffusion model.

Although ID-Animator and ConsisID have made progress in preserving identity, they overlook controllability of facial motions within the video sequence. As a result, the generated expressions are often monotonous, fail to respond flexibly to text prompts, or even duplicate the static face from the reference image across frames. The root cause is that the face embedding extracted by the Face Encoder inevitably entangles identity-irrelevant information—such as expressions and actions. When these irrelevant attributes are fed into the diffusion model, they interfere with the text prompt’s guidance over details, ultimately limiting both the diversity and accuracy of facial movements.

To tackle this problem, we propose the Identity-Action Decoupling for video generation (IaD) framework. IaD introduces two key innovations: 1) Further feature disentanglement within the face embedding to extract a purely identity-related feature while stripping away expression and motion components that are irrelevant to identity. 2) The disentangled non-identity features are modulated by the text prompt and fed into the diffusion model, thereby enabling temporally fine-grained and controllable dynamic driving of facial expressions and overall video content.

To realize these goals, we design two new loss functions: 1) Identity Decoupling Loss: enforces the purity of the identity embedding during disentanglement, preventing motion information from leaking back into the identity embedding. 2) Text Alignment Loss: encourages an one-to-one correspondence between the Facial-Action embedding and the text prompt, ensuring the generated video responds accurately over time to the prompt’s descriptions of actions and emotions. 

With the IaD framework, the model not only faithfully inherits the identity feature from the reference image, but also exhibits rich, controllable facial movements and scene variations throughout the entire video. In summary, our principal contributions are as follows:
\begin{itemize}
    \item We introduce the IaD framework, which achieves fine-grained, text-driven control over facial expressions and actions while preserving cross-temporal identity consistency. The framework is inference-friendly, requiring no additional fine-tuning for new identities.
    
    \item We design two novel losses: Identity Decoupling Loss, which purifies the identity embedding by removing identity-irrelevant information, and Text Alignment Loss, which aligns the Facial-Action embedding with textual semantics to ensure that the generated video responds precisely to the prompts.
    
    \item We conduct extensive data analyses and experiments that validate the scientific validity and effectiveness of IaD. All code, model weights, and experimental scripts will be released to facilitate reproduction and further research by the community.
\end{itemize}

\section{Related Work}
\subsection{Identity-Preserving Image Generation }
In the field of Identity‑Preserving Image Generation, early methods such as Textual Inversion \cite{gal2022image}, DreamBooth \cite{ruiz2023dreambooth}, and LoRA \cite{hu2022lora} have already achieved notable progress in customizing outputs for specific IDs \cite{bao2018towards,xiao2022identity,peng2024portraitbooth,chen2023disenbooth}. However, these approaches require the model to be fine‑tuned for every new ID, which is time‑consuming, hampers real‑time performance, and makes large‑scale deployment difficult. Recent research is breaking free from test‑time training, shifting the core challenge to how to flexibly and efficiently extract and fuse identity features \cite{kim2024instantfamily}. IP‑Adapter \cite{ye2023ip} uses CLIP \cite{radford2021learning} to extract facial image features and introduces a disentangled cross‑attention mechanism to fuse text and image representations. Nevertheless, the CLIP encoder’s discriminative power for identity‑preservation tasks is limited, so facial‑detail fidelity remains constrained. InstantID \cite{wang2024instantid} is the first to incorporate a pre‑trained ArcFace \cite{deng2019arcface} encoder to capture facial features and, together with IdentityNet, explicitly encodes reference‑face features, significantly enhancing identity consistency. Subsequent work commonly adopts a framework that pairs an ArcFace encoder (focusing on facial regions) with a CLIP encoder (covering broader visual semantics) to obtain richer and more fine‑grained identity representations \cite{chen2024dreamidentity,wu2024infinite}.
\subsection{Identity-Preserving Video Generation}
Compared with image generation, the core challenge of Identity‑Preserving Video Generation is two‑fold: maintaining identity consistency across time while achieving precise control over facial motions \cite{guo2024real,he2026video}. Early work such as Still‑moving \cite{chefer2024still} made an initial exploration without relying on any video data; DreamVideo \cite{wei2024dreamvideo} split the customization process into subject learning and motion learning; MoVideo \cite{liang2024movideo} went further by proposing explicit motion modeling and utilization strategies. However, these methods still require test‑time training, continuing the paradigms of Textual Inversion, DreamBooth, and LoRA. To reduce the high cost at inference, ID‑Animator \cite{he2024id} was the first to introduce and open‑source a “zero‑training” framework that can be used out of the box. ConsisID \cite{yuan2024identity} then improved identity stability by injecting the high‑ and low‑frequency facial information into different layers of DiT. Even so, the above studies focus too heavily on identity consistency and pay insufficient attention to the potential of text prompts for controlling facial expressions and movements, thereby limiting the diversity of the generated videos in terms of expressions and actions.

\section{Preliminary}
Before introducing our method, we briefly review video diffusion models \cite{ho2020denoising,song2020denoising}. A diffusion model learns the data distribution through a forward noise-adding process and a reverse denoising process. Given a clean image $\mathbf{x}_0 \in \mathbb{R}^{H \times W \times C}$, the forward process gradually injects Gaussian noise over $T$ steps.

\begin{equation}
q(\mathbf{x}_t \mid \mathbf{x}_{t-1})=\mathcal{N}\!\bigl(\mathbf{x}_t;\sqrt{1-\beta_t}\,\mathbf{x}_{t-1},\;\beta_t\mathbf I\bigr),\quad t=1,\dots,T
\label{equation1}
\end{equation}

Let $\alpha_t = 1 - \beta_t$ and $\bar\alpha_t = \prod_{i=1}^{t} \alpha_i$, we can get:
\begin{equation}
q(\mathbf{x}_t \mid \mathbf{x}_0)=\mathcal{N}\!\bigl(\mathbf{x}_t;\sqrt{\bar\alpha_t}\,\mathbf{x}_0,\;(1-\bar\alpha_t)\mathbf I\bigr)
\label{equation2}
\end{equation}

The objective of the reverse process is to learn $p_\theta(\mathbf{x}_{t-1}\mid\mathbf{x}_t)$ so that it approximates the true posterior $q(\mathbf{x}_{t-1}\mid\mathbf{x}_t,\mathbf{x}_0)$. It can be written as:
\begin{equation}
p_\theta(\mathbf x_{t-1}\mid\mathbf x_t)
=\mathcal N\!\bigl(\mathbf x_{t-1};\mu_\theta(\mathbf x_t,t),\sigma_t^2\mathbf I\bigr)
\label{equation3}
\end{equation}

The training objective of the model is:
\begin{equation}
\mathcal{L}_{\text{img}}
=\mathbb E_{\mathbf x_0,t,\varepsilon}\bigl[
\lVert\varepsilon-
\varepsilon_\theta(\sqrt{\bar\alpha_t}\mathbf x_0+\sqrt{1-\bar\alpha_t}\varepsilon,t)\rVert_2^2
\bigr]
% ,\quad \varepsilon\sim\mathcal N(0,\mathbf I)
\label{equation4}
\end{equation}

Extending to video diffusion models \cite{ho2022video}, a video sample is represented as $\,\mathbf v_0 \in \mathbb{R}^{F \times H \times W \times C}\,$, where $F$ denotes the number of frames. The forward and reverse distributions are transferred verbatim to the spatio-temporal domain:
\begin{equation}
q(\mathbf v_t\mid\mathbf v_{0})
=\mathcal{N}\!\bigl(\mathbf{v}_t;\sqrt{\bar\alpha_t}\,\mathbf{v}_0,\;(1-\bar\alpha_t)\mathbf I\bigr)
\label{equation5}
\end{equation}

\begin{equation}
p_\theta(\mathbf v_{t-1}\mid\mathbf v_t)
=\mathcal N\!\bigl(\mathbf v_{t-1};\mu_\theta(\mathbf v_t,t),\sigma_t^2\mathbf I\bigr).
\label{equationRe}
\end{equation}

The training objective of the model is:
\begin{equation}
\mathcal L_{\text{video}}
=\mathbb E_{\mathbf v_0,t,\varepsilon}\bigl[
\lVert\varepsilon-
\varepsilon_\theta(\sqrt{\bar\alpha_t}\mathbf v_0+\sqrt{1-\bar\alpha_t}\varepsilon,t)\rVert_2^2
\bigr]
% ,\quad \varepsilon\sim\mathcal N(0,\mathbf I)
\label{equation6}
\end{equation}

In video generation work, MM-DiT \cite{esser2024scaling} is the most mainstream backbone.

\section{Method}
% ---- figure ----  
\begin{figure*}[t]
  \centering
  % \fbox{\rule{0pt}{2in} \rule{0.97\linewidth}{0pt}}
    \includegraphics[width=0.9\linewidth]{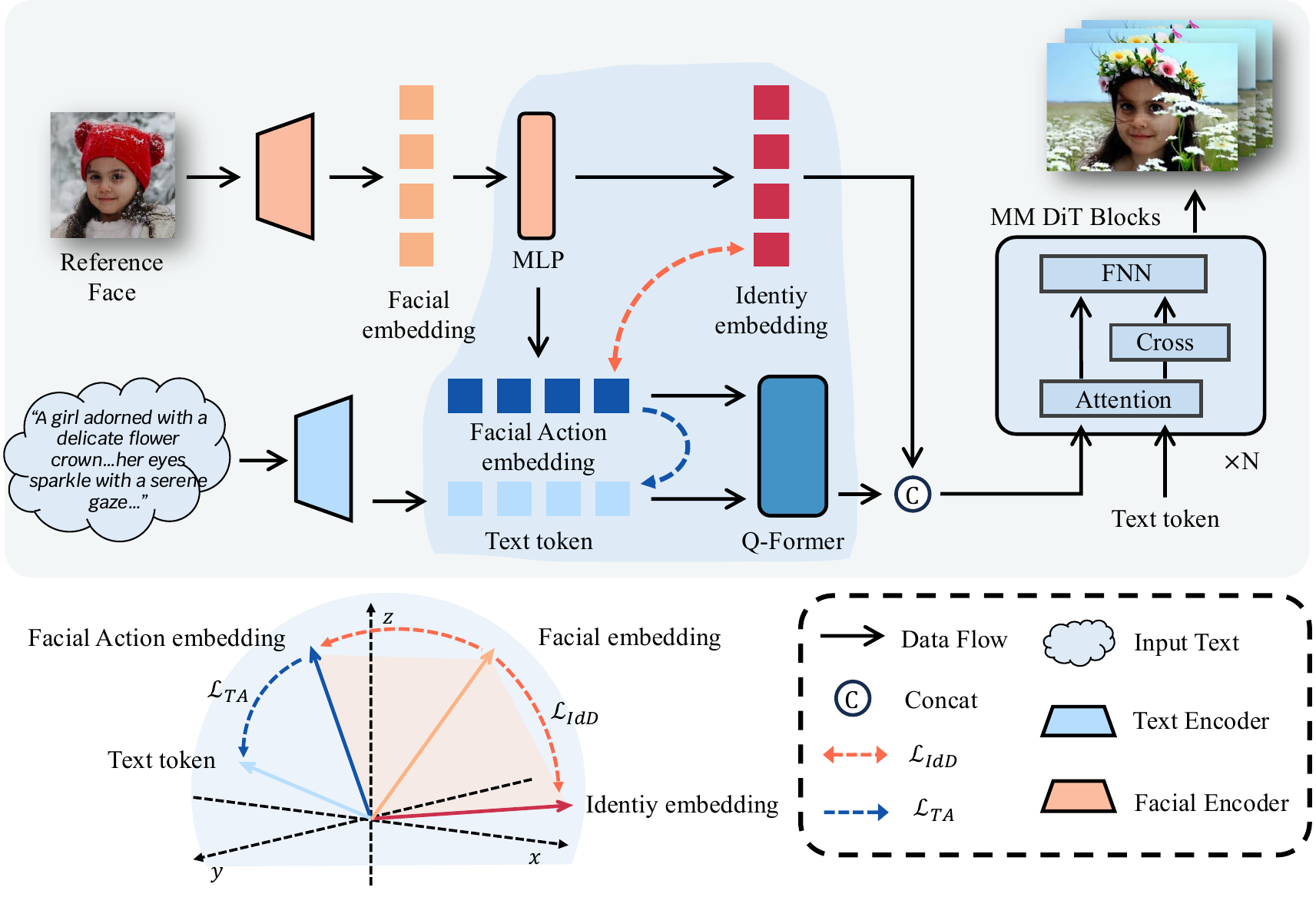}
   \caption{The framework of IaD. By decoupling the Facial embedding into an Identity embedding and a Facial-Action embedding, IaD is able to synthesize videos that preserve identity consistency while allowing precise control over facial dynamics.}
   \label{fig:framework}
\end{figure*}
% ---- ---- ---- ---- ----

\subsection{Overview}
Figure \ref{fig:framework} illustrates the framework of IaD. Given a reference face, IaD can generate a video that conforms to textual description. Concretely, a Facial Encoder first extracts a Facial embedding from the reference image; the MLP then splits this embedding into an Identity embedding and a Facial-Action embedding. The Identity embedding retains only identity information, whereas the Facial-Action embedding encodes dynamic attributes such as expressions and motions that are independent of identity. We fuse the Facial-Action embedding with the text tokens, concatenate the result with the Identity embedding, and feed the combined representation into the MM-DiT \cite{esser2024scaling} blocks to synthesize the final video sequence.

As fore Facial encoder, we couple an ArcFace encoder, which focuses on local facial details, with a CLIP encoder that captures broader visual semantics. This hybrid scheme precisely captures fine-grained facial features while providing a rich identity representation. However, it also injects considerable identity-irrelevant information into the facial embedding; if passed directly to the diffusion model, these extraneous cues would dilute the identity signal and weaken the prompt’s control. Consequently, we impose two targets: 

\begin{itemize}
\item The Identity embedding and Facial-Action embedding, obtained by splitting the Facial embedding, should be as orthogonal as possible so that they do not interfere with each other. 

\item The Facial-Action embedding must remain highly responsive to text prompts, enabling precise expression and motion control simply by editing the input text.
\end{itemize}

Therefore, we design two novel losses: Identity Decoupling Loss and Text Alignment Loss.

\subsection{Identity Decoupling Loss}
Given a reference image $x_{img}$, the Facial embedding is denoted as $e^{face}\in\mathbb{R}^{N_q\times 2D}$.

\begin{equation}
e^{face} =  \mathcal{E}_{Face}(x_{img})
\label{equation7}
\end{equation}

Here, $\mathcal{E}_{Face}$ denotes the Facial Encoder, $N_q$ is the number of query tokens and $2D$ is the dimension of  $e^{face}$. The MLP is applied to $e^{{face}}$ to split it into the identity embedding  
$e^{{id}}_k \in \mathbb{R}^{N_q \times D}$
and the facial-action embedding  
$e^{{FA}}_k \in \mathbb{R}^{N_q \times D}$.

To ensure that the Facial-Action embedding contains only identity-irrelevant information, we enforce orthogonality between the two representations in the embedding space:

\begin{equation}
\langle \hat e^{id}_{q},\;\hat e^{FA}_{q}\rangle \;=\;0,
\quad q=1\dots N_q
\label{equation8}
\end{equation}

In this step, all vectors are $L_2$ normalized. This normalization process serves to constrain the vectors to a unit sphere, which helps to prevent overfitting by reducing the complexity of the model. 

\begin{equation}
\hat e \;=\; \frac{e}{\|e\|_2+\varepsilon}\quad
\label{equation9}
\end{equation}

We compute the Identity Decoupling Loss by taking the absolute dot product between the two normalized embeddings for every token in every batch sample, then averaging the result:

\begin{equation}
\mathcal L_{IdD}
= \frac1{B N_q}\sum_{b=1}^{B}\sum_{q=1}^{N_q}
\bigl(\, \langle\,\hat{\mathbf e}^{{id}}_{b,q},\,
\hat{\mathbf e}^{{FA}}_{b,q}\rangle
\bigr)^2
\label{equation11}
\end{equation}

where $B$ is the batch size. When $\mathcal{L}_{{IdD}}\rightarrow 0$, the two subspaces become nearly orthogonal, $e^{{id}}$ and $e^{{FA}}$ are effectively disentangled.

\subsection{Text Alignment Loss}
After disentangling the Facial-Action embedding from the Identity embedding, we want the Facial-Action embedding to respond sensitively and controllably to text prompts. To achieve this, we adopt a contrastive-learning scheme: for each sample, we pull its Facial-Action embedding toward its own text representation while pushing it away from the text representations of other samples, thereby establishing a Facial-Action to text alignment.

Given $e^{{FA}}_k$, we first average its $N_q$ query tokens and then apply $L_2$ normalization:

\begin{equation}
z_i=\tfrac1{N_q}\sum_q {e^{FA}}_{iq},\hat z_i=\frac{z_i}{\lVert z_i\rVert_2+\varepsilon}
\label{equation12}
\end{equation}

For each text prompt, we feed it into a text encoder to obtain text token, project them to a $D$-dimensional space, then average over the sequence length $T$ and apply $L_2$ normalization. Q-Former \cite{li2023blip} is a commonly used lightweight module for bridging different modalities. We use it here to integrate textual information with $e^{{FA}}_k$.

\begin{equation}
c_i=\tfrac1{T}\sum_t \mathcal{E}_{text}({prompt}_{it}), 
\hat c_i=\frac{c_i}{\lVert c_i\rVert_2+\varepsilon}
\label{equation13}
\end{equation}

According to \cite{oord2018representation}, we get Equation \ref{equation14}, where $\tau$ is the temperature coefficient.
\begin{equation}
\mathcal L_{TA}
= \frac1B\sum_{i=1}^{B}
\Bigl[-\log
\frac{\exp(\langle z_i, c_i\rangle/\tau)}
{\sum_{j=1}^{B} \exp(\langle z_i, c_j\rangle/\tau)}
\Bigr]
\label{equation14}
\end{equation}

\subsection{Training Strategy}
To avert convergence issues such as gradient explosion or vanishing gradients, we adopt a staged, progressive training strategy. The core idea is to preserve the discriminative power of the pre-trained modules as much as possible in the early phase, thereby providing a stable optimization baseline for subsequent end-to-end fine-tuning.

All parameters of Text Encoder and Facial Encoder are frozen throughout training; they act solely as feature extractors, retaining the robust representations learned from large-scale text and face corpora. First, we freeze the MM-DiT blocks (denoted $D_{\mathrm{MM}}$) and update only the MLP parameters $\Theta_{\mathrm{MLP}}$ and the Q-Former parameters $\Theta_{\mathrm{Q}}$ as shown in Equation \ref{equation15}.

\begin{equation}
\min_{\Theta_{\mathrm{MLP}},\,\Theta_{\mathrm{Q}}}\;
\mathcal{L}\bigl(\mathcal{E}_{face},\,\mathcal{E}_{text},\,D_{\mathrm{MM}};\,\Theta_{\mathrm{MLP}},\,\Theta_{\mathrm{Q}}\bigr)
\label{equation15}
\end{equation}

where $\mathcal{L}$ is the total loss function. This phase lets the MLP and Q-Former learn the initial cross-modal disentanglement and fusion mechanisms without interference from the high-dimensional transformer blocks. Once $\mathcal{L}$ plateaus at a stable threshold, we unfreeze $D_{\mathrm{MM}}$ and optimize the entire parameter set $\Theta = \{\Theta_{D_{\mathrm{MM}}},\,\Theta_{\mathrm{MLP}},\,\Theta_{\mathrm{Q}}\}$ jointly:

\begin{equation}
\min_{\Theta}\;
\mathcal{L}\bigl(\mathcal{E}_{face},\,\mathcal{E}_{text};\Theta)
    \label{equation16}
\end{equation}

With global back-propagation now enabled, the network can fine-tune the synergy among the modality-specific sub-networks, leading to deeper feature complementarity and stronger task alignment.
% % ---- figure ----  
\begin{figure}[h]
  \centering
  % \fbox{\rule{0pt}{2in} \rule{0.97\linewidth}{0pt}}
    \includegraphics[width=1\linewidth]{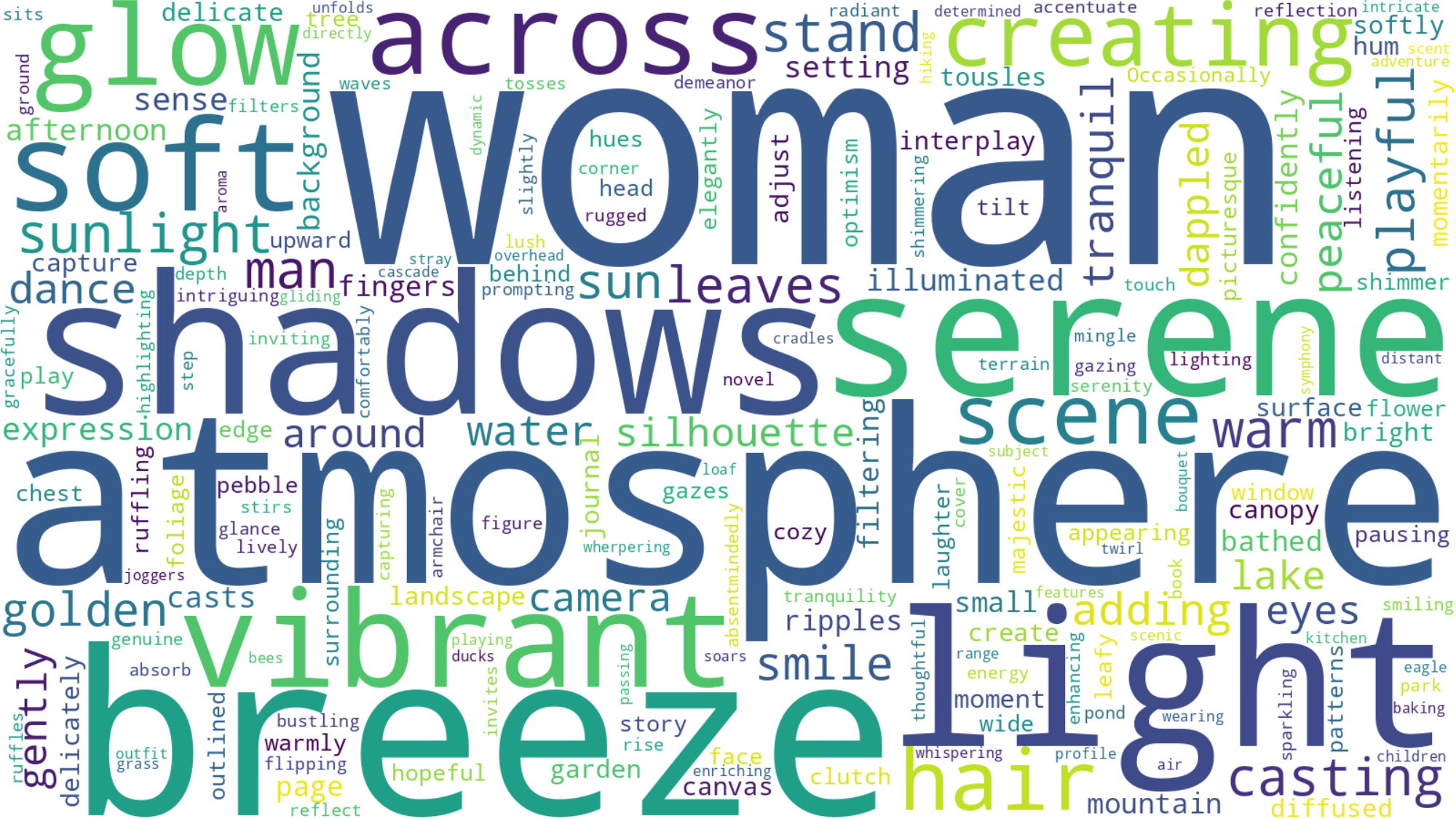}
   \caption{Word cloud of prompts.}
   \label{fig:wordcloud}
\end{figure}
% % ---- ---- ---- ---- ----

% ---- figure ----  
\begin{figure*}[h]
  \centering
  % \fbox{\rule{0pt}{2in} \rule{0.97\linewidth}{0pt}}
    \includegraphics[width=0.99\linewidth]{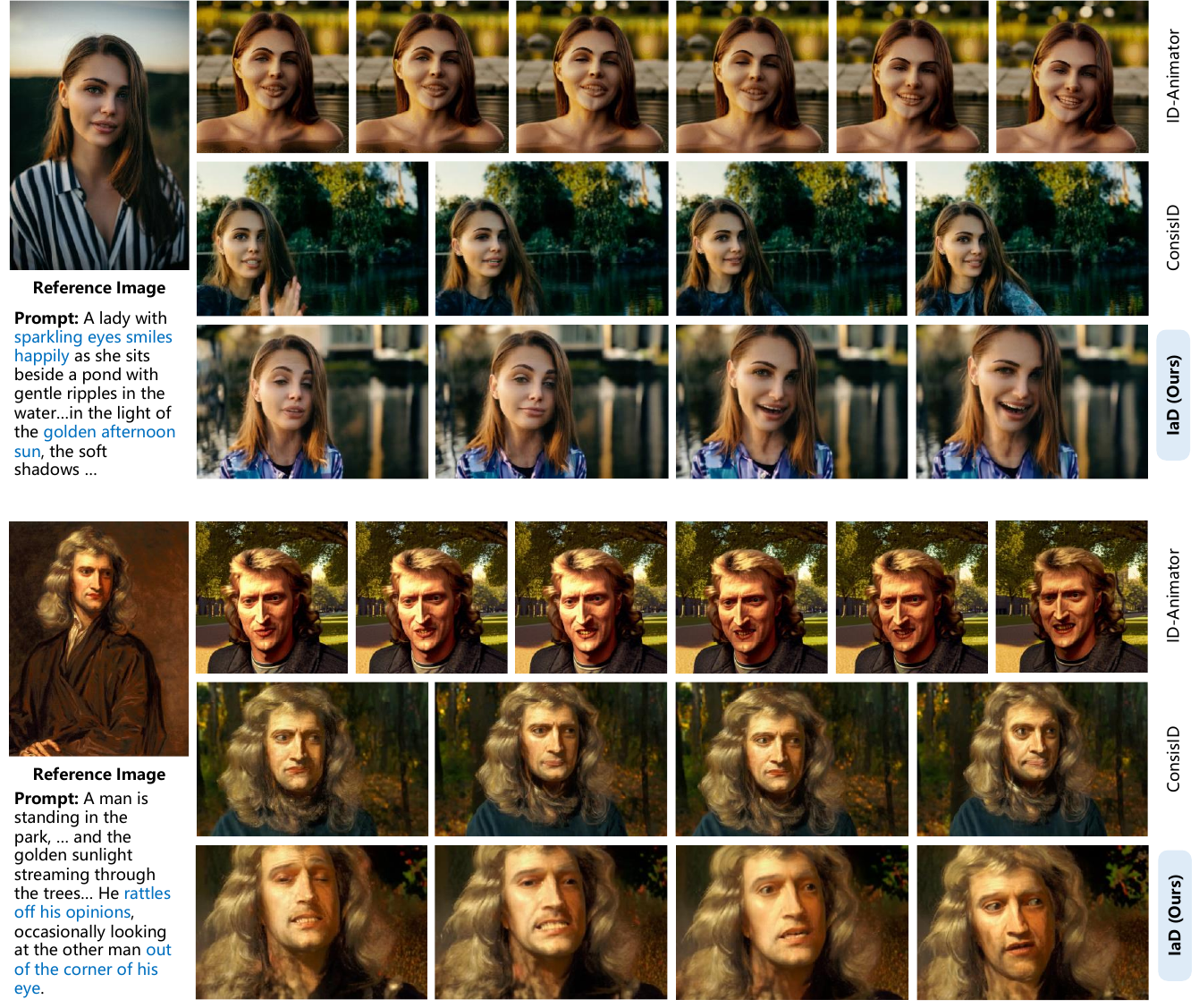}
   \caption{Visual comparison with ID-Animator and ConsisID. Compared with ID-Animator and ConsisID, IaD generates videos with richer facial motions and expressions while exhibiting closer adherence to the prompts.}
   \label{fig:comparison}
\end{figure*}
% ---- ---- ---- ---- ----

\section{Experiment}

\subsection{Experiment Setup}
\textbf{Implementation details.} We adopt CogVideoX-5B \cite{yang2024cogvideox} as the baseline model. During the first training stage, the learning rate is set to $1 \times 10^{-6}$ and is raised to $3 \times 10^{-6}$ in the second stage. The temperature coefficient in Equation \ref{equation14} is fixed at $\tau = 0.7$. Optimization is performed with AdamW, whose hyper-parameters are $\beta_1 = 0.9$ and $\beta_2 = 0.95$, together with a cosine-annealing scheduler with restarts. Concretely, we apply a warm-up of $500$ steps before the main training phase. During gradient updates, we clip the norm to $\lVert g \rVert_2 \le 1.0$ to enhance stability. Training is conducted in bf16 mixed precision, reducing memory consumption and accelerating computation. Following the pipeline provided by ConsisID, we construct a training set comprising 47,868 samples.

% ---- ---- ---- ---- ----
\begin{table*}[h]
\centering
\caption{Quantitive comparison ID-Animator and ConsisID . Higher FaceSim-Arc, FaceSim-Cur and CLIPScore values indicate better result, while lower FID scores signify better quality. \textbf{Bold indicates the best performance.}}  
\scalebox{1}{
\begin{tabular}{ccccc}
% >{\columncolor[HTML]{BDD7EE}}c }

\toprule
            & FaceSim-Arc $\uparrow$ & FaceSim-Cur $\uparrow$ & CLIPScore $\uparrow$ & FID $\downarrow$   \\ \midrule
ID-Animator & 0.31        & 0.34        & 24.13     & 129.26 \\ 
ConsisID    & 0.57        & 0.54        & 27.15     & 144.35 \\ 
\rowcolor[HTML]{BDD7EE} 
IaD (Ours)  & \textbf{0.67}        & \textbf{0.71}        & \textbf{28.75}     & \textbf{114.67} \\ \bottomrule
\end{tabular}
}
\label{tab:main}
\end{table*}
% ---- ---- ---- ---- ----

\subsection{Comparison With Existing Methods}
To verify the effectiveness of IaD, we compare it with the state-of-the-art open-source methods ID-Animator and ConsisID. Specifically, we select 50 images that are not included in the training set and create 60 diverse, fine-grained prompts for evaluation. Identity preservation is assessed with FaceSim-Arc \cite{deng2019arcface} and FaceSim-Cur \cite{yuan2024identity}, whereas video quality is measured using FID \cite{heusel2017gans} and CLIPScore \cite{hessel2021clipscore}.

\noindent \textbf{Quantitative Comparison.} As shown in Table \ref{tab:main}, IaD outperforms all existing baselines across the four objective metrics. Its FaceSim-Arc reaches $0.67$, marking an improvement of roughly $116\%$ over ID-Animator ($0.31$) and $17.5\%$ over ConsisID ($0.57$). Likewise, FaceSim-Cur rises to $0.71$, up from $0.34$ (ID-Animator) and $0.54$ (ConsisID), corresponding to gains of approximately $108\%$ and $31\%$, respectively—highlighting superior identity preservation. CLIPScore climbs to $28.75$, exceeding the scores of ID-Animator ($24.13$) and ConsisID ($27.15$) by $19.2\%$ and $5.9\%$, indicating tighter semantic alignment between the textual prompt and the generated sequence. Meanwhile, FID drops to $114.67$, representing reductions of $11.3\%$ and $20.6\%$ relative to ID-Animator ($129.26$) and ConsisID ($144.35$), evidencing markedly improved visual realism. We attribute these advantages to the proposed identity–action disentanglement architecture, which simultaneously preserves identity features and flexibly controls facial dynamics, yielding balanced and significant gains in identity consistency, semantic relevance, and overall visual quality.

\noindent \textbf{Qualitative Comparisons.} As illustrated in Figure \ref{fig:comparison}, IaD shows a clear advantage over ID-Animator and ConsisID on the identity-preserving video-generation task.

First, IaD adheres more faithfully to the textual prompt. In the female-subject example, the prompt stresses “sparkling eyes” and “a joyful smile.” IaD’s output strictly follows these cues, maintaining the smile while reproducing fine-grained details such as eye highlights and subtle expression changes. In the male-subject example, the prompt notes that “he occasionally glances sideways while speaking.” IaD generates a natural facial performance that matches this description, accurately reproducing the sideways glance—demonstrating its ability to capture and realize complex, prompt-specified facial behaviors.

Second, IaD produces a richer spectrum of facial motions and expressions. For the female subject, the expression evolves from a gentle smile to more animated dynamics, conveying nuanced emotional shifts that the competing methods fail to deliver. For the male subject, IaD not only renders the sideways glance but also additional details such as mouth movements, resulting in a more lifelike and expressive video, whereas the baselines exhibit limited diversity and subtlety in facial motion.

In summary, IaD preserves identity while more precisely following prompt instructions and generating videos with diverse, fine-grained facial actions and expressions, underscoring its superiority in identity-preserving video generation.

\subsection{Ablation Study}

\textbf{Ablation on $\mathcal L_{IdD}$ and $\mathcal L_{TA}$.} As shown in Table \ref{tab:ablation}, when both losses are disabled, the model’s identity preservation, semantic alignment, and visual quality are all at their lowest (FaceSim-Arc/Cur only $0.50/0.48$, CLIPScore $23.73$, FID $126.65$).  

Activating only the identity-disentanglement loss $\mathcal L_{IdD}$ raises FaceSim-Arc to $0.56$ and FaceSim-Cur to $0.52$, confirming its effectiveness in strengthening identity features; however, without semantic constraints, CLIPScore increases only to $25.33$, and FID even climbs to $138.64$, indicating some degradation of the image distribution.  

Conversely, using only the text–action alignment loss $\mathcal L_{TA}$ yields a slight CLIPScore gain to $25.56$, yet identity metrics fall (FaceSim-Arc $0.52$) and FID worsens to $140.97$, showing that better semantic alignment alone cannot secure identity consistency or overall realism.  

When $\mathcal L_{IdD}$ and $\mathcal L_{TA}$ are combined, all four metrics reach their best values: FaceSim-Arc/Cur rise to $\mathbf{0.67}/\mathbf{0.71}$, CLIPScore climbs to $\mathbf{28.75}$, and FID drops to $\mathbf{114.67}$. These results demonstrate that the two losses are complementary: $\mathcal L_{IdD}$ consolidates identity and image details, while $\mathcal L_{TA}$ reinforces text-driven action semantics. Working together, they simultaneously enhance identity consistency, semantic relevance, and visual realism, markedly improving the overall quality of the generated sequences.

% ---- ---- ---- ---- ----
\begin{table*}[h]
\centering
\caption{Ablation on $\mathcal L_{IdD}$ and $\mathcal L_{TA}$. \textbf{Bold indicates the best performance.}} 
\scalebox{1}{
\begin{tabular}{cccccc}
\toprule
$\mathcal L_{IdD}$   & $\mathcal L_{TA}$   & FaceSim-Arc $\uparrow$ & FaceSim-Cur $\uparrow$ & CLIPScore $\uparrow$ & FID  $\downarrow$   \\ \midrule
\xmark  & \xmark  & 0.50        & 0.48        & 23.73     & 126.65 \\
\cmark  & \xmark  & 0.56        & 0.52        & 25.33     & 138.64 \\
\xmark  & \cmark  & 0.52        & 0.53        & 25.56     & 140.97 \\
\rowcolor[HTML]{BDD7EE} 
\cmark  & \cmark  & \textbf{0.67}        & \textbf{0.71}        & \textbf{28.75}     & \textbf{114.67} \\ \bottomrule
\end{tabular}
}
\label{tab:ablation}
\end{table*}
% ---- ---- ---- ---- ----

\begin{figure}[h]
  \centering
  % \fbox{\rule{0pt}{2in} \rule{0.6\linewidth}{0pt}}
    \includegraphics[width=1\linewidth]{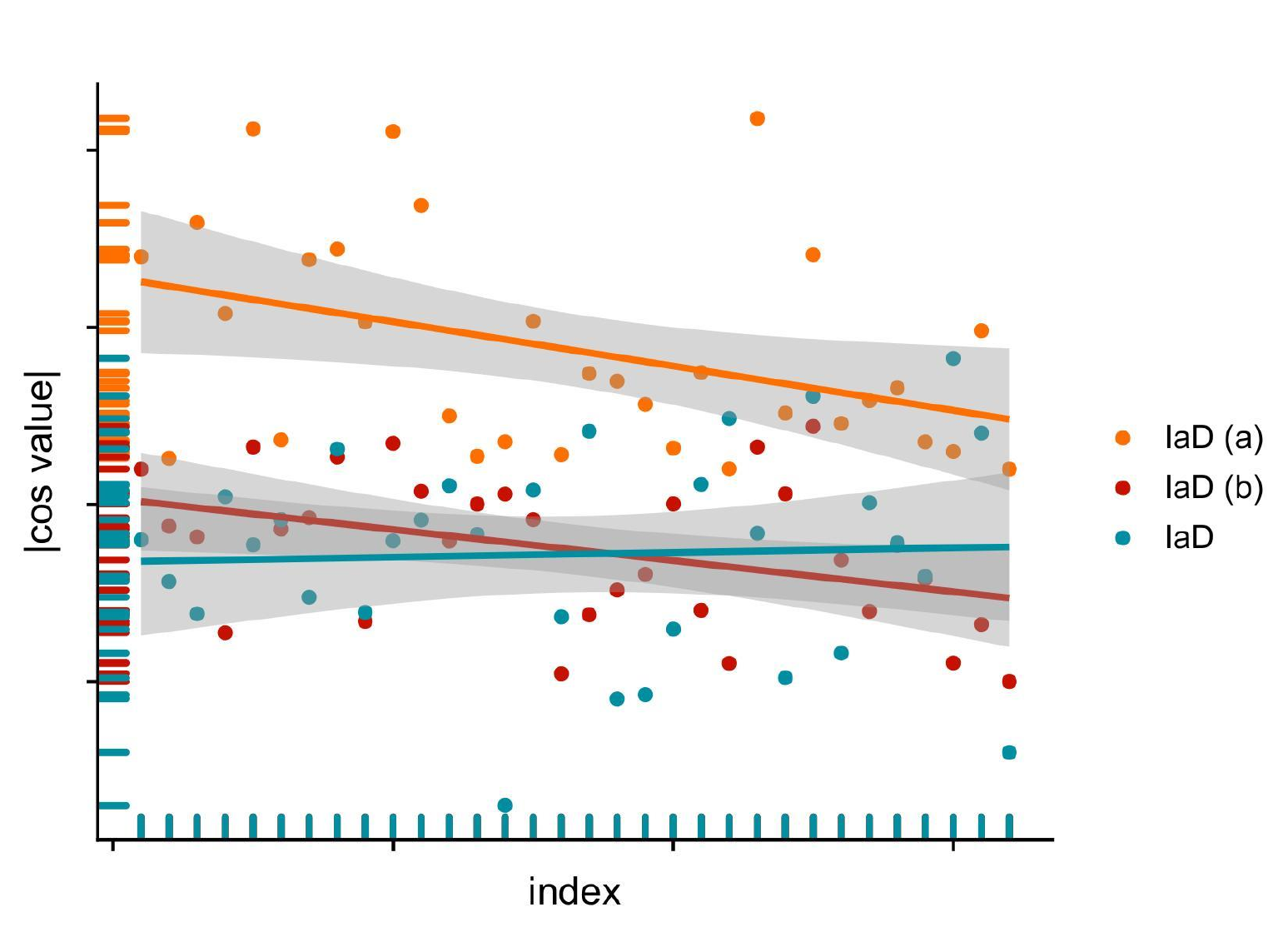}
   \caption{The cosine similarity value between Identity embedding and Facial-Action embedding, the horizontal axis represents $N_q$ index. IaD (a) represents training without $\mathcal L_{IdD}$, IaD (b) represents training without $\mathcal L_{TA}$, and IaD represents training with both.}
   \label{fig:visual}
\end{figure}

\noindent \textbf{Cosine-similarity Comparison.} Figure \ref{fig:visual} presents the cosine-similarity values between the Identity embedding and the Facial-Action embedding under three training variants. By comparing IaD(a),IaD(b) and IaD, we can clearly observe the respective contributions of the two loss functions $\mathcal L_{IdD}$ and $\mathcal L_{TA}$.  

Under the IaD(a) condition, $\mathcal{L}_{IdD}$ was not introduced during training, resulting in consistently high cosine similarity values. This indicates a high correlation between Identity embedding and Facial-Action embedding, suggesting that using $\mathcal{L}_{TA}$ alone struggles to effectively distinguish identity-related features. In contrast, under the IaD(b) condition, $\mathcal{L}_{TA}$ was excluded during training, leading to a significant reduction in cosine similarity values compared to IaD(a). This demonstrates that $\mathcal{L}_{IdD}$ effectively enables Identity embedding and Facial-Action embedding to achieve orthogonality in the feature space, thereby separating identity-related features.

 Under the IaD condition, both $\mathcal{L}_{IdD}$ and $\mathcal{L}_{TA}$ were combined during training. In this case, the cosine similarity values under IaD(b) training conditions closely align with those of IaD, indicating that the addition of $\mathcal{L}_{TA}$ did not significantly improve cosine similarity. From the shaded regions in the figure, it can be observed that the scatter points for IaD and IaD(b) are relatively concentrated, suggesting stable results. In contrast, the scatter points for IaD(a) are more dispersed, indicating greater fluctuations in cosine similarity values when $\mathcal{L}_{IdD}$ is not used.

In summary, the $\mathcal{L}_{IdD}$ loss function plays a pivotal role in decoupling identity information from action information. Moreover, the $\mathcal{L}_{IdD}$ and $\mathcal{L}_{TA}$ loss functions are not conflicting in their optimization objectives and can be compatibly integrated into the training process to jointly enhance model performance.
% ---- figure ----  
% ---- ---- ---- ---- ----

\section{Conclusion}
We have introduced Identity–Action Decoupling (IaD), a unified framework for identity-preserving text-to-video generation that explicitly separates static identity cues from dynamic action cues in latent space. The proposed Identity-Decoupling Loss $\mathcal L_{IdD}$ enforces cross-temporal identity consistency, while the Text-Alignment Loss $\mathcal L_{TA}$ guides motion and expression toward the semantics of the input prompt.  Working together, the two losses resolve the long-standing conflict between faithfully maintaining a subject’s appearance and producing vivid, prompt-driven facial dynamics—achieving superior identity fidelity, semantic alignment and visual realism over prior approaches such as ID-Animator and ConsisID, all without any subject-specific fine-tuning. Qualitative comparisons show that IaD prevents identity drift when expressions change, and ablation studies confirm that $\mathcal L_{IdD}$ and $\mathcal L_{TA}$ are complementary: disabling either term noticeably harms both perceptual quality and prompt adherence, whereas their combination produces the most coherent results.  

While IaD substantially improves the controllability–fidelity trade-off, certain challenges remain, including handling extreme poses, long video durations and multi-person scenarios.  Future work will explore further to enhance temporal consistency and expressive richness.  We believe the decoupling paradigm put forward in this paper offers a solid foundation for the next generation of personalized, prompt-driven video synthesis systems.

\bibliography{aaai2026}

@article{chefer2024still,
  title={Still-moving: Customized video generation without customized video data},
  author={Chefer, Hila and Zada, Shiran and Paiss, Roni and Ephrat, Ariel and Tov, Omer and Rubinstein, Michael and Wolf, Lior and Dekel, Tali and Michaeli, Tomer and Mosseri, Inbar},
  journal={ACM Transactions on Graphics (TOG)},
  volume={43},
  number={6},
  pages={1--11},
  year={2024},
  publisher={ACM New York, NY, USA}
}

@article{yuan2024identity,
  title={Identity-Preserving Text-to-Video Generation by Frequency Decomposition},
  author={Yuan, Shenghai and Huang, Jinfa and He, Xianyi and Ge, Yunyuan and Shi, Yujun and Chen, Liuhan and Luo, Jiebo and Yuan, Li},
  journal={arXiv preprint arXiv:2411.17440},
  year={2024}
}

@article{kim2025subject,
  title={Subject-driven Video Generation via Disentangled Identity and Motion},
  author={Kim, Daneul and Zhang, Jingxu and Jin, Wonjoon and Cho, Sunghyun and Dai, Qi and Park, Jaesik and Luo, Chong},
  journal={arXiv preprint arXiv:2504.17816},
  year={2025}
}

@article{he2024id,
  title={Id-animator: Zero-shot identity-preserving human video generation},
  author={He, Xuanhua and Liu, Quande and Qian, Shengju and Wang, Xin and Hu, Tao and Cao, Ke and Yan, Keyu and Zhang, Jie},
  journal={arXiv preprint arXiv:2404.15275},
  year={2024}
}

@article{song2024idprotector,
  title={IDProtector: An Adversarial Noise Encoder to Protect Against ID-Preserving Image Generation},
  author={Song, Yiren and Yang, Pei and Ci, Hai and Shou, Mike Zheng},
  journal={arXiv preprint arXiv:2412.11638},
  year={2024}
}

@article{li2024personalvideo,
  title={PersonalVideo: High ID-Fidelity Video Customization without Dynamic and Semantic Degradation},
  author={Li, Hengjia and Qiu, Haonan and Zhang, Shiwei and Wang, Xiang and Wei, Yujie and Li, Zekun and Zhang, Yingya and Wu, Boxi and Cai, Deng},
  journal={arXiv preprint arXiv:2411.17048},
  year={2024}
}

@article{wang2024lia,
  title={LIA: Latent Image Animator},
  author={Wang, Yaohui and Yang, Di and Bremond, Francois and Dantcheva, Antitza},
  journal={IEEE Transactions on Pattern Analysis and Machine Intelligence},
  year={2024},
  publisher={IEEE}
}

@article{qiu2024moviecharacter,
  title={MovieCharacter: A Tuning-Free Framework for Controllable Character Video Synthesis},
  author={Qiu, Di and Chen, Zheng and Wang, Rui and Fan, Mingyuan and Yu, Changqian and Huang, Junshi and Wen, Xiang},
  journal={arXiv preprint arXiv:2410.20974},
  year={2024}
}

@article{hu2022lora,
  title={Lora: Low-rank adaptation of large language models.},
  author={Hu, Edward J and Shen, Yelong and Wallis, Phillip and Allen-Zhu, Zeyuan and Li, Yuanzhi and Wang, Shean and Wang, Lu and Chen, Weizhu and others},
  journal={ICLR},
  volume={1},
  number={2},
  pages={3},
  year={2022}
}

@article{gal2022image,
  title={An image is worth one word: Personalizing text-to-image generation using textual inversion},
  author={Gal, Rinon and Alaluf, Yuval and Atzmon, Yuval and Patashnik, Or and Bermano, Amit H and Chechik, Gal and Cohen-Or, Daniel},
  journal={arXiv preprint arXiv:2208.01618},
  year={2022}
}

@article{fei2025ingredients,
  title={Ingredients: Blending Custom Photos with Video Diffusion Transformers},
  author={Fei, Zhengcong and Li, Debang and Qiu, Di and Yu, Changqian and Fan, Mingyuan},
  journal={arXiv preprint arXiv:2501.01790},
  year={2025}
}

@article{deng2025cinema,
  title={CINEMA: Coherent Multi-Subject Video Generation via MLLM-Based Guidance},
  author={Deng, Yufan and Guo, Xun and Wang, Yizhi and Fang, Jacob Zhiyuan and Wang, Angtian and Yuan, Shenghai and Yang, Yiding and Liu, Bo and Huang, Haibin and Ma, Chongyang},
  journal={arXiv preprint arXiv:2503.10391},
  year={2025}
}

@inproceedings{ruiz2023dreambooth,
  title={Dreambooth: Fine tuning text-to-image diffusion models for subject-driven generation},
  author={Ruiz, Nataniel and Li, Yuanzhen and Jampani, Varun and Pritch, Yael and Rubinstein, Michael and Aberman, Kfir},
  booktitle={Proceedings of the IEEE/CVF conference on computer vision and pattern recognition},
  pages={22500--22510},
  year={2023}
}

@inproceedings{bao2018towards,
  title={Towards open-set identity preserving face synthesis},
  author={Bao, Jianmin and Chen, Dong and Wen, Fang and Li, Houqiang and Hua, Gang},
  booktitle={Proceedings of the IEEE conference on computer vision and pattern recognition},
  pages={6713--6722},
  year={2018}
}

@inproceedings{xiao2022identity,
  title={Identity preserving loss for learned image compression},
  author={Xiao, Jiuhong and Aggarwal, Lavisha and Banerjee, Prithviraj and Aggarwal, Manoj and Medioni, Gerard},
  booktitle={Proceedings of the IEEE/CVF Conference on Computer Vision and Pattern Recognition},
  pages={517--526},
  year={2022}
}

@article{wang2024instantid,
  title={Instantid: Zero-shot identity-preserving generation in seconds},
  author={Wang, Qixun and Bai, Xu and Wang, Haofan and Qin, Zekui and Chen, Anthony and Li, Huaxia and Tang, Xu and Hu, Yao},
  journal={arXiv preprint arXiv:2401.07519},
  year={2024}
}

@inproceedings{peng2024portraitbooth,
  title={Portraitbooth: A versatile portrait model for fast identity-preserved personalization},
  author={Peng, Xu and Zhu, Junwei and Jiang, Boyuan and Tai, Ying and Luo, Donghao and Zhang, Jiangning and Lin, Wei and Jin, Taisong and Wang, Chengjie and Ji, Rongrong},
  booktitle={Proceedings of the IEEE/CVF Conference on Computer Vision and Pattern Recognition},
  pages={27080--27090},
  year={2024}
}

@article{chen2023disenbooth,
  title={Disenbooth: Identity-preserving disentangled tuning for subject-driven text-to-image generation},
  author={Chen, Hong and Zhang, Yipeng and Wu, Simin and Wang, Xin and Duan, Xuguang and Zhou, Yuwei and Zhu, Wenwu},
  journal={arXiv preprint arXiv:2305.03374},
  year={2023}
}

@article{kim2024instantfamily,
  title={Instantfamily: Masked attention for zero-shot multi-id image generation},
  author={Kim, Chanran and Lee, Jeongin and Joung, Shichang and Kim, Bongmo and Baek, Yeul-Min},
  journal={arXiv preprint arXiv:2404.19427},
  year={2024}
}

@article{ye2023ip,
  title={Ip-adapter: Text compatible image prompt adapter for text-to-image diffusion models},
  author={Ye, Hu and Zhang, Jun and Liu, Sibo and Han, Xiao and Yang, Wei},
  journal={arXiv preprint arXiv:2308.06721},
  year={2023}
}

@inproceedings{radford2021learning,
  title={Learning transferable visual models from natural language supervision},
  author={Radford, Alec and Kim, Jong Wook and Hallacy, Chris and Ramesh, Aditya and Goh, Gabriel and Agarwal, Sandhini and Sastry, Girish and Askell, Amanda and Mishkin, Pamela and Clark, Jack and others},
  booktitle={International conference on machine learning},
  pages={8748--8763},
  year={2021},
  organization={PmLR}
}

@inproceedings{deng2019arcface,
  title={Arcface: Additive angular margin loss for deep face recognition},
  author={Deng, Jiankang and Guo, Jia and Xue, Niannan and Zafeiriou, Stefanos},
  booktitle={Proceedings of the IEEE/CVF conference on computer vision and pattern recognition},
  pages={4690--4699},
  year={2019}
}

@inproceedings{chen2024dreamidentity,
  title={DreamIdentity: enhanced editability for efficient face-identity preserved image generation},
  author={Chen, Zhuowei and Fang, Shancheng and Liu, Wei and He, Qian and Huang, Mengqi and Mao, Zhendong},
  booktitle={Proceedings of the AAAI Conference on Artificial Intelligence},
  volume={38},
  number={2},
  pages={1281--1289},
  year={2024}
}

@inproceedings{wu2024infinite,
  title={Infinite-ID: Identity-preserved Personalization via ID-semantics Decoupling Paradigm},
  author={Wu, Yi and Li, Ziqiang and Zheng, Heliang and Wang, Chaoyue and Li, Bin},
  booktitle={European Conference on Computer Vision},
  pages={279--296},
  year={2024},
  organization={Springer}
}

@inproceedings{wei2024dreamvideo,
  title={Dreamvideo: Composing your dream videos with customized subject and motion},
  author={Wei, Yujie and Zhang, Shiwei and Qing, Zhiwu and Yuan, Hangjie and Liu, Zhiheng and Liu, Yu and Zhang, Yingya and Zhou, Jingren and Shan, Hongming},
  booktitle={Proceedings of the IEEE/CVF Conference on Computer Vision and Pattern Recognition},
  pages={6537--6549},
  year={2024}
}

@inproceedings{liang2024movideo,
  title={Movideo: Motion-aware video generation with diffusion model},
  author={Liang, Jingyun and Fan, Yuchen and Zhang, Kai and Timofte, Radu and Van Gool, Luc and Ranjan, Rakesh},
  booktitle={European Conference on Computer Vision},
  pages={56--74},
  year={2024},
  organization={Springer}
}

@article{ho2022video,
  title={Video diffusion models},
  author={Ho, Jonathan and Salimans, Tim and Gritsenko, Alexey and Chan, William and Norouzi, Mohammad and Fleet, David J},
  journal={Advances in Neural Information Processing Systems},
  volume={35},
  pages={8633--8646},
  year={2022}
}

@article{ho2020denoising,
  title={Denoising diffusion probabilistic models},
  author={Ho, Jonathan and Jain, Ajay and Abbeel, Pieter},
  journal={Advances in neural information processing systems},
  volume={33},
  pages={6840--6851},
  year={2020}
}

@article{song2020denoising,
  title={Denoising diffusion implicit models},
  author={Song, Jiaming and Meng, Chenlin and Ermon, Stefano},
  journal={arXiv preprint arXiv:2010.02502},
  year={2020}
}

@inproceedings{esser2024scaling,
  title={Scaling rectified flow transformers for high-resolution image synthesis},
  author={Esser, Patrick and Kulal, Sumith and Blattmann, Andreas and Entezari, Rahim and M{\"u}ller, Jonas and Saini, Harry and Levi, Yam and Lorenz, Dominik and Sauer, Axel and Boesel, Frederic and others},
  booktitle={Forty-first international conference on machine learning},
  year={2024}
}

@article{oord2018representation,
  title={Representation learning with contrastive predictive coding},
  author={Oord, Aaron van den and Li, Yazhe and Vinyals, Oriol},
  journal={arXiv preprint arXiv:1807.03748},
  year={2018}
}

@inproceedings{li2023blip,
  title={Blip-2: Bootstrapping language-image pre-training with frozen image encoders and large language models},
  author={Li, Junnan and Li, Dongxu and Savarese, Silvio and Hoi, Steven},
  booktitle={International conference on machine learning},
  pages={19730--19742},
  year={2023},
  organization={PMLR}
}

@article{yang2024cogvideox,
  title={CogVideoX: Text-to-Video Diffusion Models with An Expert Transformer},
  author={Yang, Zhuoyi and Teng, Jiayan and Zheng, Wendi and Ding, Ming and Huang, Shiyu and Xu, Jiazheng and Yang, Yuanming and Hong, Wenyi and Zhang, Xiaohan and Feng, Guanyu and others},
  journal={arXiv preprint arXiv:2408.06072},
  year={2024}
}

@article{heusel2017gans,
  title={Gans trained by a two time-scale update rule converge to a local nash equilibrium},
  author={Heusel, Martin and Ramsauer, Hubert and Unterthiner, Thomas and Nessler, Bernhard and Hochreiter, Sepp},
  journal={Advances in neural information processing systems},
  volume={30},
  year={2017}
}

@article{hessel2021clipscore,
  title={Clipscore: A reference-free evaluation metric for image captioning},
  author={Hessel, Jack and Holtzman, Ari and Forbes, Maxwell and Bras, Ronan Le and Choi, Yejin},
  journal={arXiv preprint arXiv:2104.08718},
  year={2021}
}

@article{tao2024lcgen,
  title={Lcgen: Mining in low-certainty generation for view-consistent text-to-3d},
  author={Tao, Zeng and Yang, Tong and Lin, Junxiong and Mai, Xinji and Wang, Haoran and Wang, Beining and Zhou, Enyu and Wang, Yan and Zhang, Wenqiang},
  journal={Advances in Neural Information Processing Systems},
  volume={37},
  pages={20276--20303},
  year={2024}
}

@article{mai2026agentic,
  title={Agentic RL scaling law: Spontaneous code execution for mathematical problem solving},
  author={Mai, Xinji and Xu, Haotian and Wang, Weinong and Zhang, Yingying and Zhang, Wenqiang and others},
  journal={Advances in Neural Information Processing Systems},
  volume={38},
  pages={7325--7340},
  year={2026}
}

@article{mai2025cues,
  title={CuES: A Curiosity-driven and Environment-grounded Synthesis Framework for Agentic RL},
  author={Mai, Shinji and Zhai, Yunpeng and Chen, Ziqian and Chen, Cheng and Zou, Anni and Tao, Shuchang and Liu, Zhaoyang and Ding, Bolin},
  journal={arXiv preprint arXiv:2512.01311},
  year={2025}
}

@inproceedings{wang2026hi,
  title={Hi-ef: Benchmarking emotion forecasting in human-interaction},
  author={Wang, Haoran and Mai, Xinji and Tao, Zeng and Lin, Junxiong and Tong, Xuan and Pan, Ivy and Yan, Shaoqi and Wang, Yan and Gao, Shuyong},
  booktitle={Proceedings of the AAAI Conference on Artificial Intelligence},
  volume={40},
  number={3},
  pages={2110--2118},
  year={2026}
}

@article{he2026video,
  title={Video Generation Models as World Models: Efficient Paradigms, Architectures and Algorithms},
  author={He, Muyang and Guo, Hanzhong and Lin, Junxiong and Yu, Yizhou},
  journal={arXiv preprint arXiv:2603.28489},
  year={2026}
}

@article{guo2024real,
  title={Real-time identity defenses against malicious personalization of diffusion models},
  author={Guo, Hanzhong and Nie, Shen and Du, Chao and Pang, Tianyu and Sun, Hao and Li, Chongxuan},
  journal={arXiv preprint arXiv:2412.09844},
  year={2024}
}

@inproceedings{tong2025component,
  title={Component-aware Unsupervised Logical Anomaly Generation for Industrial Anomaly Detection},
  author={Tong, Xuan and Chang, Yang and Zhao, Qing and Yu, Jiawen and Wang, Boyang and Lin, Junxiong and Lin, Yuxuan and Mai, Xinji and Wang, Haoran and Tao, Zeng and others},
  booktitle={2025 IEEE International Conference on Robotics and Automation (ICRA)},
  pages={16722--16729},
  year={2025},
  organization={IEEE}
}


\end{document}